\documentclass{article}

\usepackage[final]{neurips_2024}

\usepackage[utf8]{inputenc} % allow utf-8 input
\usepackage[T1]{fontenc}    % use 8-bit T1 fonts
\usepackage{hyperref}       % hyperlinks
\usepackage{url}            % simple URL typesetting
\usepackage{booktabs}       % professional-quality tables
\usepackage{amsfonts}       % blackboard math symbols
\usepackage{nicefrac}       % compact symbols for 1/2, etc.
\usepackage{microtype}      % microtypography
\usepackage{xcolor}         % colors
\setcitestyle{square}
\usepackage{graphicx} 
\usepackage{todonotes}
\usepackage{wrapfig}
\usepackage{float}
\title{A Deep Generative Model for the Design of Synthesizable Ionizable Lipids}

\author{%
  Yuxuan Ou\\
  Department of Engineering\\
  University of Cambridge\\
  Cambridge, CB2 1PZ \\
  \texttt{yuxuan.ou@trinity.ox.ac.uk} \\
  \And
  Jingyi Zhao\\
  Department of Engineering\\
  University of Cambridge\\
  Cambridge, CB2 1PZ \\
  \texttt{jz610@cam.ac.uk}\\
  \And
  Austin Tripp\\
  Department of Engineering\\
  University of Cambridge\\
  Cambridge, CB2 1PZ \\
  \texttt{ajt212@cam.ac.uk} \\
  \And
  Morteza Rasoulianboroujeni\\
  School of Pharmacy\\
  University of Wisconsin-Madison\\
  Madison, WI 53706 \\
  \texttt{rasoulianbor@wisc.edu} \\
  \And
  José Miguel Hernández-Lobato\\
  Department of Engineering\\
  University of Cambridge\\
  Cambridge, CB2 1PZ \\
  \texttt{jmh233@cam.ac.uk} \\
}

\begin{document}

\maketitle

\begin{abstract}
Lipid nanoparticles (LNPs) are vital in modern biomedicine, enabling the effective delivery of mRNA for vaccines and therapies by protecting it from rapid degradation. Among the components of LNPs, ionizable lipids play a key role in RNA protection and facilitate its delivery into the cytoplasm. However, designing ionizable lipids is complex. Deep generative models can accelerate this process and explore a larger candidate space compared to traditional methods. Due to the structural differences between lipids and small molecules, existing generative models used for small molecule generation are unsuitable for lipid generation. To address this, we developed a deep generative model specifically tailored for the discovery of ionizable lipids. Our model generates novel ionizable lipid structures and provides synthesis paths using synthetically accessible building blocks, addressing synthesizability. This advancement holds promise for streamlining the development of lipid-based delivery systems, potentially accelerating the deployment of new therapeutic agents, including mRNA vaccines and gene therapies.
\end{abstract}

% {
% \color{blue}
% Austin's suggested outline.

\section{Introduction}
\begin{figure}[!ht]
    \centering
    \includegraphics[width=0.3\linewidth]{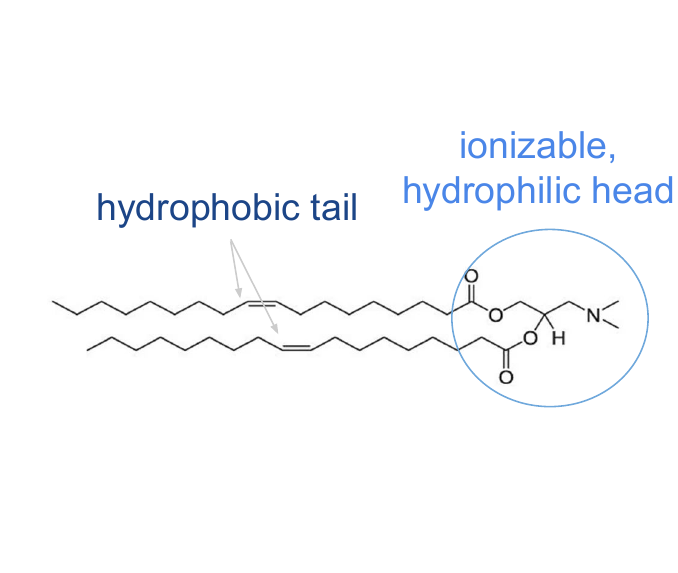}
    \caption{\textbf{An Example of Ionizable Lipid Structure.}}
    \label{fig:lipidexamples}
\end{figure}

% 1.5 page (contain a LNP and a ionizable lipid figure)
% Messenger RNA (mRNA) is a powerful tool in biomedicine, used in genetic information transfer, developing viral vaccines, protein replacement therapies, and cancer immunotherapies \citep{mrna1, mrna2}. However, naked mRNA is unstable and easily degraded by nucleases, which underscores the need for effective delivery mechanisms \citep{mrna3}. 
The successful development of COVID vaccine paved the way for the clinical application of lipid nanoparticles (LNPs)  to deliver different kinds of messenger RNA(mRNA) \citep{covidlnp, mrna4}. The standard structure of LNPs includes four main components: ionizable lipids, cholesterol, helper lipids, and PEGylated lipids \citep{mrna5}. Ionizable lipids are critical as they condense the negatively charged mRNA during LNP formulation and help it escape from endosomes and enter the cytoplasm of target cells to express the protein of interest \citep{Agile,mrna6}.
An ionizable lipid is an amphiphillic molecule with an ionizable, hydrophilic head and several hydrophobic tails \citep{mrna7}. An example structure of an ionizable lipid is shown in Figure \ref{fig:lipidexamples}.  
% The ionizable head determines the lipid’s charge and polarity which is crucial for complexation with mRNA \citep{mrna8}.

Designing ionizable lipids is time-consuming and labor-intensive. Combinatorial chemistry provides a fast and cost-effective method to produce these lipids in large quantities, which allows researchers perform high-throughput screenings of ionizable lipids \citep{ il2}. For example, by using a Ugi-based three-component reaction (3-CR), researchers can rapidly generate a library of 1,080 ionizable lipids \citep{agile2}. This collection can assist in identifying an ionizable lipid for the application of interest such as activation of stimulator of interferon genes (STING), which is useful for delivering mRNA vaccines. However,
despite these advances, the method has limitations. The diversity of ionizable lipids generated remains constrained due to the restricted range of lipid heads and tails available.  Therefore, designing and testing a broader and more diverse range of lipids remains challenging.

Machine learning methods, particularly generative models, provide a solution for efficiently exploring vast molecular search spaces \citep{ml1, ml6, ml7, ml3, synthesisdags}. These models capture the distributions of existing molecules, learn how molecular structures correlate with their physical properties, and use this knowledge to predict new molecules. Various generative models have been proposed for molecule design, capable of generating new molecules through either molecular graphs or string representations \citep{ml4}. Despite significant advances in machine learning for drug discovery, previous works have shown that generative models are prone to producing molecules which chemists find very difficult to synthesize \citep{synthesisdags}.

Existing research in machine learning for LNP design primarily focuses on predicting LNP transfection efficiency \citep{il2, agile2, ding2023machinelearningguidedlipidnanoparticle, 10.1093/bioinformatics/btae342}. In contrast, our work addresses the generation of ionizable lipids. Specifically, we adapt an existing approach, Synthesis-DAGs \citep{synthesisdags}, which simultaneously generates molecules and their synthesis routes, to the task of generating synthesizable ionizable lipids \citep{ml5}.

Our main constributions are as follows: 
\begin{itemize}
    \item Extract a broad range of synthetically accessible building blocks for ionizable lipid and construct an ionizable lipid synthesis dataset for training the generator.
    \item Develop a generator capable of producing ionizable lipids along with their synthesis pathways from the building blocks.
    \item Iteratively fine-tune the generator to identify optimal ionizable lipid structures with synthesis pathways for high mRNA transfection efficiency in HeLa cells.
\end{itemize}

% Essentially say why you want to generate lipids, and that this paper proposes 2 methods to do it.

\section{Background}
In this section, we first introduce the generator on which our work is based, Synthesis-DAGs, then we introduce the mRNA transfection efficiency prediction model used in the optimization process.
\subsection{Synthesis-DAGs}
Synthesis-DAGs is a general molecule generator that our method builds upon \citep{syndag}. It can generate molecules along with their synthesis pathways. Below, we briefly introduce how Synthesis-DAGs work.%
% \todo[author=Austin]{Does everything below need to be a subsubsection? The headings take up a lot of space. I might use the \texttt{paragraph} command instead.}

\paragraph{Representing Synthesis Paths as DAGs} 
Synthesis routes are represented as directed acyclic graphs (DAGs). Examples are shown in Figure \ref{fig:generateexamples}. Building blocks are displayed in blue boxes. These building blocks undergo chemical reactions to form intermediate products, shown in light purple boxes. Finally, the final product is generated, shown in dark purple boxes. 
\paragraph{Serializing the Construction of DAGs as Action Sequences}The DAGs are serialized into action sequences by defining three classes of actions: \textit{Node-addition}, \textit{Building block molecular identity}, and \textit{Connectivity choice}. Every molecule in the DAG is considered a node and is classified as either a building block node or a product node. The \textit{Node-addition} action determines which type of node to add. \textit{Building block molecular identity} specifies which molecule to choose from the building block pool. \textit{Connectivity choice} decides which nodes to connect to the next product and whether that product is an intermediate product or final product.
\paragraph{Modeling the Probability Distribution of the Action Sequence} A shared RNN models the distribution of the action sequence. At each step, the RNN computes a context vector, which is passed into a feedforward action network to predict the next action. There are three action networks in total, one for each action class. In order to feed the action sequence into the RNN model, the actions are converted into action embeddings. The detailed action types, action choices and action embeddings are listed in Table \ref{tab:dag_generation}.
\begin{table}[htbp]
    \centering
    \renewcommand{\arraystretch}{1.5} % Increases row height
    \setlength{\tabcolsep}{5pt} % Reduces column padding to fit within margins
    \begin{tabular}{|p{2.5cm}|p{5cm}|p{5cm}|} % Adjusts column width
        \hline
        \textbf{Action Type} & \textbf{Action} & \textbf{Action Embedding} \\ 
        \hline
        Node-addition             & Building block node              & Learnable vector $h_B$ \\ 
        \cline{2-3}
                             & Product node                     & Learnable vector $h_P$ \\ 
        \hline
        Building block molecular identity & Select a molecule in the building block pool & Building block’s molecular embedding \\ 
        \hline
        Connectivity choice  & Select a reactant node            & Reactant’s molecular embedding \\ 
        \cline{2-3}
                             & Identify as an intermediate product (continue) & Learnable vector $h_I$ \\ 
        \cline{2-3}
                             & Identify as a final product (stop) & Learnable vector $h_F$ \\ 
        \hline
    \end{tabular}
     \vspace{0.5cm} % Adds vertical space before the caption
    \caption{\textbf{Action Type, Action, and Action Embeddings of DAG Generation}}
    \label{tab:dag_generation}
\end{table}

\paragraph{Reaction Predictor} During sampling, a reaction predictor is used to perform reaction prediction at the product nodes. In Synthesis-DAGs, Molecular Transformer is used for this task \citep{MolecularTransformer}.

In this work, we focus on adapting Bradshaw’s Synthesis-DAGs model for use as an ionizable lipid DAG generator. The limitations of directly applying the original method to lipid generation and the detailed adaptations we made are discussed in Section 5.

\subsection{AGILE framework} When optimizing ionizable lipids for high mRNA transfection efficiency, we use the AGILE model \citep{Agile} to predict the mRNA transfection efficiency of our generated ionizable lipids. The AGILE model is a deep learning framework designed to predict the transfection efficiency of ionizable lipids in specific cells. It is part of the AI-Guided Ionizable Lipid Engineering (AGILE) platform. The training process of the prediction model occurs in two stages. First, a graph encoder is pre-trained on a virtual library of 60,000 chemically diverse lipids using contrastive learning. In the second stage, the model is fine-tuned with wet-lab mRNA transfection efficiency data. We use the model fine-tuned on data with mRNA transfection efficiency in HeLa cells as labels to make predictions. 

\section{Lipid Property Predictors}
Our method relies on both a lipid classifier and an ionizable lipid classifier to determine whether the generated product is a lipid or an ionizable lipid. In this section, we introduce the lipid classifier, a binary classification model that determines whether a molecule is lipid-like. We then explain how ionizable lipids are identified by examining their net charge at physiological and acidic pH.
\paragraph{Lipid Classifier} The architecture of our lipid classifier is based on the message passing neural networks framework, Chemprop \citep{chemprop}. It consists of three message passing layers to integrate molecular features, followed by two feedforward layers for property prediction. The training dataset includes 180,000 lipids and 180,000 non-lipid molecules. Lipid data are sourced from public datasets LIPID MAPS and SwissLipids, as well as synthetic data generated using graph-based structural motifs \citep{lmsd, swisslipid, graphmotif}. Non-lipid molecules are obtained from the PubChem database \citep{pubchem}. Our classifier demonstrates excellent performance, achieving both a Receiver Operating Characteristic Area Under the Curve (ROC-AUC) score and a Precision-Recall Area Under the Curve (PR-AUC) score above 0.9999.

\paragraph{Ionizable Lipid Classifier} Ionizable lipids carry a positive charge at acidic pH, enabling them to condense RNAs into LNPs, but are neutral at physiological pH to minimize toxicity \citep{ionizablelipidnetcharge}. To filter based on this criterion, we consider the net charge at pH 5 (acidic) and pH 7.4 (physiological). The net charge is calculated by first estimating the pKa values of the lipid's acidic and basic groups using MolGpka \citep{molgpka}, followed by applying the Henderson-Hasselbalch equation to determine the lipid's net charge at both pH values.

\paragraph{Property Predictors Validation} 
To further validate our lipid property predictors, we curated a dataset of over 2,500 ionizable lipids sourced from previously published studies \citep{il2,https://doi.org/10.1002/anie.202310401,Liu2021-sq, https://doi.org/10.1002/adfm.202303795, https://doi.org/10.1002/adhm.201901487, Wei2023-uh}. Importantly, this dataset is entirely independent of the training data used for our lipid classifier. Using this dataset, we evaluated the predictors' performance, with the lipid classifier achieving a high accuracy of 98.32\%. Additionally, the ionizability predictor demonstrated exceptional performance, accurately classifying all ionizable lipids in the dataset.
\section{Dataset Construction}
% 1 page
In this section, we first explain how we select synthetically accessible ionizable lipid building blocks by detailing the filtering criteria used to choose lipid heads and tails from the ZINC20 dataset of synthetically accessible components. After forming a building block pool, we demonstrate how we synthesize ionizable lipids based on these building blocks and construct the ionizable lipid synthesis dataset.
\subsection{Building Block Extraction}
% 1/2 page
We begin the construction of an ionizable lipid synthesis dataset by creating a building block set of ionizable lipid heads and tails. To ensure that all building blocks are synthetically accessible, we select them from the ZINC20 dataset \citep{zinc20}, a publicly available database that includes commercially available compounds. 
We apply three filtering criteria to identify the ionizable lipid heads. The first criterion is molecular weight; since most lipids, including heads and tails, have a molecular weight of no more than 1000 g/mol. We set our limit at no more than 500 g/mol. The second criterion is about solubility preference, selecting compounds with a LogP value less than zero, where LogP represents the log of the partition coefficient between octanol, a hydrophobic oil, and water. The third criterion focus on charge characteristics; we seek ionizable heads that maintain a near-zero charge at physiological pH and become positively charged in acidic environments. Compounds containing a nitrogen atom in their structure are likely to meet this ionization requirement. Thus, the third criterion involves selecting molecules that contain amine groups. By applying these criteria, we identify a set of 2.7 million ionizable head molecules.

As for selecting lipid tails, we aim to ensure that the molecule structurally resembles a lipid tail, and the tail set is diverse. We initially use LipidAnalyzer (a toolbox already implemented) to extract tails from the LIPID MAPS\citep{lmsd}. LIPID MAPS contains 48,548 unique lipid structures. From these, we extract a total of 8,176 unique lipid tails. To ensure that all the lipid tails are synthetically accessible, we find similar lipid tail molecules in the ZINC dataset based on these extracted tails. The search tool we use called `cartblanche' \citep{cartblanche2024}. We measure similarity by calculating the
graph edit distance (GED) between two molecules, which involves finding the minimum
number of graph operations needed to transform one graph into another. The second search criterion is that at least one of the molecule’s Tanimoto similarity coefficients must be greater than 0.5, using either Daylight fingerprints or ECFP4 fingerprints.
After identifying all synthetically accessible components, we filter them by checking whether the tail is reactive; this is determined by the presence of a heteroatom. Through this search, we identify a total of 15,302 synthetically accessible lipid tails.
\subsection{Lipid Synthesis Dataset Construction}
% 1/2 page
After obtaining the synthetically accessible lipid heads and tails, our next step is to generate ionizable lipids with multiple tails. In this study, we focus on lipids with 1-3 tails. We start by identifying lipid heads that are most likely to react with the selected lipid tails. An analysis of the lipid tails revealed the functional groups with high occurrences, among which the top three functional groups that can react with those in the tails are the carboxylic acid group, amine group, and hydroxyl group. Consequently, we further refine our selection of ionizable heads based on a combination of these three functional groups. Specifically, we require the lipid head to contain 1-3 functional groups to facilitate chemical reactions that attach the tails. This count of 1-3 functional groups excludes the amine group used to ensure ionizability.  

We synthesize ionizable lipids by sequentially adding lipid tails to lipid heads. The chemical reactions that combine a lipid tail with a lipid head or an intermediate product are simulated using Chemformer \citep{Chemformer}, a reaction prediction model. After the final tail addition, the product is formed. We first apply the lipid classifier to determine if it is a lipid. If classified as a lipid, we then apply the ionizable lipid classifier to assess its ionizability.

\section{Synthesizable Ionizable Lipid Generator}
\begin{figure}[!ht]
    \centering
    \includegraphics[width=0.8\linewidth]{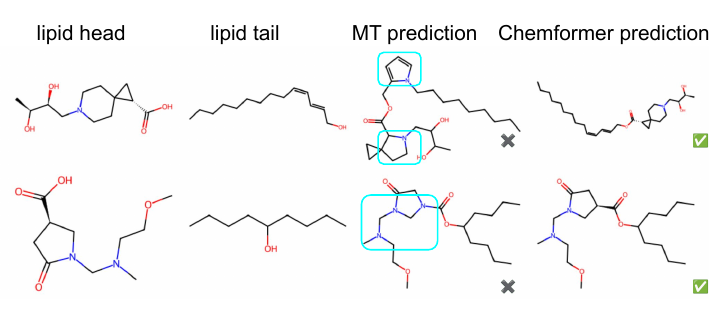}
    \caption{\textbf{Examples of Predictions from the Original and Improved Reaction Predictors.} The original reaction predictor, Molecular Transformer, fails to accurately predict reactions between lipid heads and tails, with errors indicated by blue boxes. In contrast, Chemformer successfully predicts the correct reactions.}
    \label{fig:rtexamples}
\end{figure}

% 1 page
% Explain the two methods. Maybe one subsection for each of them, and a subsection for dataset construction.
In this section, we explain why the original Synthesis-DAGs method cannot be directly applied to lipid generation and describe the adjustments we made to address these limitations.
\paragraph{Limitations of the Synthesis-DAGs Method in Generating Ionizable Lipids} We sampled 1,000 molecules using the original Synthesis-DAGs model trained on the USPTO reaction dataset \citep{USPTO}. Of these, only 53 were classified as lipids, and 13 as ionizable lipids. This low efficiency indicates that the original model cannot be directly applied to ionizable lipid generation. One clear reason is that the USPTO reaction dataset is not lipid-specific. Additionally, the reaction predictor in Synthesis-DAGs fails to accurately predict reactions between lipid tails and heads \citep{MolecularTransformer}. Example predictions from the Molecular Transformer are shown in Figure \ref{fig:rtexamples}. As seen, the Molecular Transformer tends to make copy-paste errors when predicting reactions between large molecules, highlighted in the blue boxes. It incorrectly alters structures that are not involved in the chemical reaction. For instance, in the first reaction, the location of the carboxylic acid and the ring structure consisting of five carbons and one nitrogen are both altered on the lipid head (shown in the bottom blue box), while a ring structure is incorrectly added to the lipid tail (shown in the top blue box).

To address these two problems, we made two adjustments to the original Synthesis-DAGs model. First, we generated an ionizable lipid synthesis dataset. Second, we used Chemformer \citep{Chemformer} as the reaction predictor, which is capable of correctly predicting reactions for larger molecules.

% In synthesis-DAGs, a directed acyclic graph(DAG) is a graph characterized by edges with directions, where "acyclic" indicates that it is impossible to start at any node and follow a continuous path that returns to the same node. In the context of chemical synthesis, each node within a DAG represents a unique molecule, and each edge delineates the direction of a reaction from reactants to products.

% The construction of a lipid synthesis DAG is accomplished through a series of defined actions. Actions are categorized into three types: adding a node, choosing a building block, and making a connectivity choice. Each action type is chosen based on the choices made in the preceding step.  The objective is to accurately predict the next action choice based on its corresponding action
% type.
\subsection{Ionizable Lipid Synthesis Dataset}
Using the method outlined in Section 4, we generated an ionizable lipid synthesis dataset to ensure the model is trained specifically in the domain of ionizable lipids. The dataset contains 70,536 synthesis paths and includes 43,741 building blocks, comprising 38,431 unique lipid heads and 5,310 unique lipid tails. We randomly split the dataset into training, validation, and test sets, with 63,480, 3,582, and 3,582 data points, respectively. The training set includes 5,905 one-tail, 21,301 two-tail, and 36,274 three-tail ionizable lipids.

\subsection{A Better Reaction Predictor}
To address the limitations of Molecular Transformer, we adopted Chemformer, a more advanced Transformer-based sequence-to-sequence model \citep{Chemformer}. Chemformer takes reactant SMILES strings as input and outputs the predicted SMILES string of the product. It is initially pre-trained on approximately 100 million SMILES strings from the ZINC15 dataset \citep{Zinc15}, followed by fine-tuning on the USPTO-MIT dataset \citep{usptomit}, which includes around 470,000 reactions.

We selected Chemformer for several key reasons. First, Chemformer is state-of-the-art for reaction prediction on the USPTO Mixed and USPTO Separated benchmarks \citep{Chem1}. Second, Chemformer significantly reduces copy-paste errors, examples are shown in Figure \ref{fig:rtexamples}. For example, in these two examples, Chemformer makes accurate predictions without altering structures not involved in the reaction.

Apart from these two adjustments, the generator’s structure, training, and sampling methods remain the same as in Synthesis-DAGs \citep{synthesisdags}.
\section{Experiments}
% 2 page (Contain figures of results)
In this section, we present the experimental results. First, we introduce the baseline methods used for comparison. Next, we describe the experiment implementation. Finally we show results and analysis.
\subsection{Baselines}
\paragraph{Random Generation}
Random generation involves selecting a lipid head and one to three lipid tails at random, followed by sequential chemical reactions using Chemformer to combine them. This method generates raw training data without filtering for lipid property predictors, as described in Section 4. The key difference between this approach and ours lies in how the building blocks are selected. We refer to this baseline as 'Random + Chem'
\paragraph{Original Synthesis-DAGs Trained on Our Ionizable Lipid Synthesis Dataset}
We mentioned in Section 5 that one of the reasons the original Synthesis-DAGs cannot be used to generate ionizable lipids is due to the training data. Now, we
compare our method (refered as DAG + Chem), which is trained on the ionizable lipid dataset and using Chemformer as reaction prediction, to the original Synthesis-DAGs model, which was trained on our
generated ionizable lipid dataset. Because this model uses DAGs to represent synthesis paths and employs the Molecular Transformer in the sampling process to predict reactions, we refer to this baseline as 'DAG+MT'. The only difference between our method and this baseline is the reaction predictor used. 
\paragraph{Synthesis List Generation: A Linear Structure to Represent the Synthesis Path} Analyzing the synthesis paths in our training dataset reveals a linear process, characterized by the following steps: \begin{itemize} \item Adding a lipid head. \item Sequentially adding lipid tails that react to generate intermediate or final products. \end{itemize} Based on this linear progression, we developed a linear synthesis pathway generator. This approach simplifies the representation of the synthesis process as a linear list, compared to DAG-based methods. The structure of the list is as follows: \begin{center} [building block node, building block node, product node, building block node, product node, ...] \end{center} Each item in the action sequence represents a molecule's index, with action embeddings corresponding to the molecule embeddings. At each product node, the context vector is input into a binary classification network to determine whether to stop the list generation. This baseline investigates the impact of different synthesis pathway representations. All other model components, including the reaction predictor (Chemformer), remain the same as in our approach. We refer to this baseline as 'List + Chem'.

\subsection{Experiment Implementations}
% metrics, baselines, experiment implementations
The ionizable lipid generator operates in two stages: training and sampling. During training, a reaction predictor is not required, but it is used in the sampling stage. The reaction predictor is hosted on a Flask server running the Chemformer model. The generator sends requests to the server, which performs inference and returns the results. Each model is trained for 10 epochs using the Adam optimizer with a learning rate of 0.0001. During sampling, we draw 100 batches from the trained model, with each batch requesting 200 samples. However, the actual number of generated products is lower due to failures in the reaction prediction model. The experiments were conducted on a Tesla P100 GPU with 16GB of memory.
\subsection{Experiment Results}
\begin{figure}[!ht]
    \centering
    \includegraphics[width=1\linewidth]{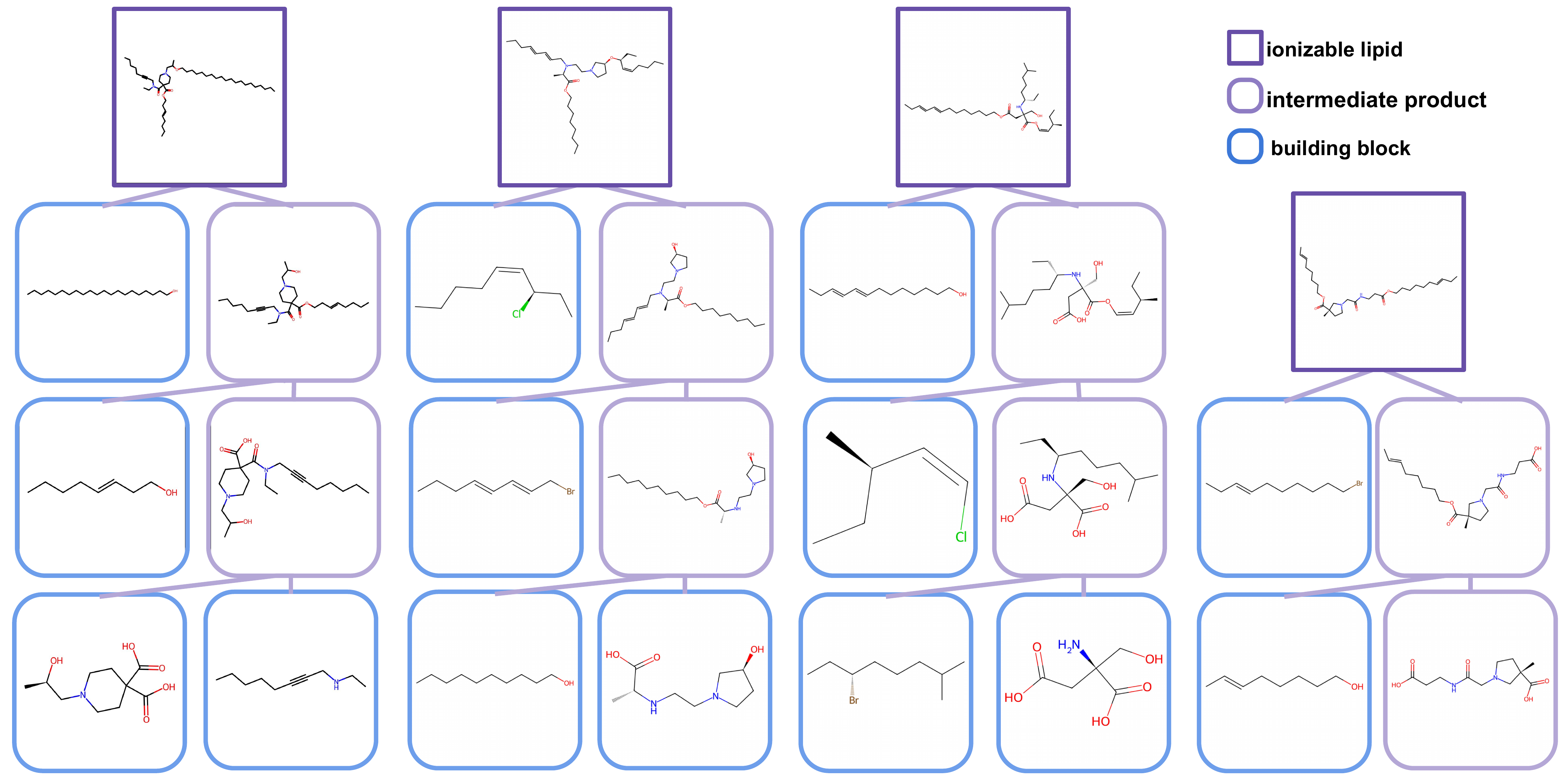}
    \caption{\textbf{Examples of Generated Ionizable Lipids and Their Synthesis Paths.} The synthesis paths show the building blocks and intermediate products. Our model can generate ionizable lipids with one to three tails.}
    \label{fig:generateexamples}
\end{figure}

We analyze the ability to generate ionizable lipids in terms of generation efficiency and quality, as shown in Table \ref{tab:result1} and Table \ref{tab:result2}, respectively. Table \ref{tab:result1} presents the ionizable lipid generation rate and lipid generation rate for different methods. In Table \ref{tab:result2}, we evaluate the validity (whether the generated SMILES can be parsed by RDKit \citep{rdkit}), uniqueness (whether the generated molecules are different from one another), novelty (whether the generated molecules different from training data), FCD (Fr\'echet ChemNet Distance, whether the generated samples have similar chemical and biological properties to those of the training data) \citep{FCD}, Synthetic Accessibility score (SA score) \citep{SAscore}. In Figure \ref{fig:generateexamples}, we present four examples of the generated ionizable lipids along with their synthesis paths.

\begin{table}[H]
\centering
\caption{\textbf{Experimental Results on the Generation Efficiency of Ionizable Lipid Generation Methods.}
 Lipid rate and ionizable lipid rate indicate the proportion of lipids and ionizable lipids in the generated samples.}
\label{tab:result1}
\begin{tabular}{@{}lcccc@{}}
\toprule
Methods & Generated Samples & Lipid Rate ($\uparrow$)& Ionizable Lipid Rate ($\uparrow$)\\ \midrule
DAG+Chem (Ours)   & 14,148 & \textbf{92.6\%} & \textbf{83.4\%} \\
List+Chem  & 15,590 & 90.6\% & 78.3\% \\
DAG+MT  & 14,575 & 72.0\% & 50.7\% \\
Random+Chem   & 309,075 & 80.9\% & 69.2\% \\
\bottomrule
\end{tabular}
\end{table}

\begin{table}[h!]
\centering
\caption{\textbf{Comparison of the quality of generated ionizable lipids.}}
\begin{tabular}{@{}lcccccc@{}}
\toprule
Methods & Validity ($\uparrow$) & Uniqueness ($\uparrow$) & Novelty ($\uparrow$) & FCD ($\downarrow$)  & SA Score ($\downarrow$) \\ \midrule
DAG+Chem (Ours)  & 1.000 & 0.999 & 0.999 & \textbf{3.797} & 4.182 \\
List+Chem & 1.000 & \textbf{1.000} & 0.999 & 4.119 & \textbf{4.013} \\
DAG+MT    & 1.000 & 0.999 & \textbf{1.000} & 4.128 & 4.141 \\ 
Random+Chem& 1.000 & 1.000 & None & None  & 4.201 \\
\bottomrule
\end{tabular}
\label{tab:result2}
\end{table}

% \subsubsection{The Impact of Training a Generative Model on Data}
% We begin by comparing the results of our method with random generation (DAG + Chem vs. Random + Chem). Our method improves the lipid rate and ionizable lipid rate by 11.7\% and 14.2\%, respectively, over random generation. This demonstrates that training a model on existing DAG data for synthesizing lipidincreases lipid generation efficiency. The improvement comes from better prediction of the next building block and action through model training.
\subsubsection{Method Performance}
As illustrated in Table~\ref{tab:result1} and Table~\ref{tab:result2}, our ionizable lipid generator performs well in terms of both the ionizable lipid generation rate and the lipid quality. The ionizable lipid rate among all generated samples reaches 83.4\%, significantly surpassing that achieved through random generation ('Random + Chem'). Regarding quality analysis, validity, uniqueness, and novelty are notably high. Furthermore, the relatively low Fréchet ChemNet Distance compared to other methods indicates that, although the ionizable lipids generated by our method differ from those in the training set, their chemical and biological properties remain within the same distribution, demonstrating successful generalization.

\subsubsection{The Impact of Reaction Predictor Performance} This comparison focuses on using Chemformer as the reaction predictor versus using Molecular Transformer ('DAG + Chem' vs. 'DAG + MT'). Our method significantly outperforms 'DAG + MT' in both lipid rate and ionizable lipid rate. The lower efficiency of Molecular Transformer as a reaction predictor has a substantial impact on the generation of ionizable lipids \citep{Chem1}.  Chemformer possesses approximately 45 million trainable parameters, which is substantially more than the 12 million parameters in Molecular Transformer. Generally, a larger number of parameters can enhance model performance due to increased learning capacity. Moreover, Chemformer undergoes a two-stage training process. It is initially pre-trained in a self-supervised manner on 100 million SMILES strings from the ZINC15 database before being fine-tuned on downstream tasks. 
In contrast, Molecular Transformer is trained directly on a downstream dataset without the intermediary step of pre-training. This direct approach may limit its ability to thoroughly learn the SMILES syntax, which is essential for predicting correct chemical structures \citep{Duan}. Thus, the use of pre-trained models, which have already developed a foundational understanding of SMILES syntax, is critical for accurate prediction of chemical products. 

\subsubsection{The Impact of Data Structure Representing Synthesis Pathway} In Table \ref{tab:result1}, the third comparison between 'DAG + Chem' and 'List + Chem' shows that the DAG generator outperforms the linear method in generating both lipids and ionizable lipids. While the uniqueness and novelty metrics are nearly identical for both methods, the DAG generator achieves a lower FCD score, indicating that the lipids produced more closely match the training distribution.
The primary reason for this is that DAG representations include diverse action embeddings, providing richer information to the RNN, which helps generate more accurate hidden states for decision-making. Additionally, DAGs can model more complex synthesis routes, particularly those involving reactions of intermediate products.

\subsection{Optimization Towards Ionizable Lipids With High Transfection Efficiency} 
\begin{figure}[htbp!]
    \centering
    \includegraphics[width=1.0\linewidth]{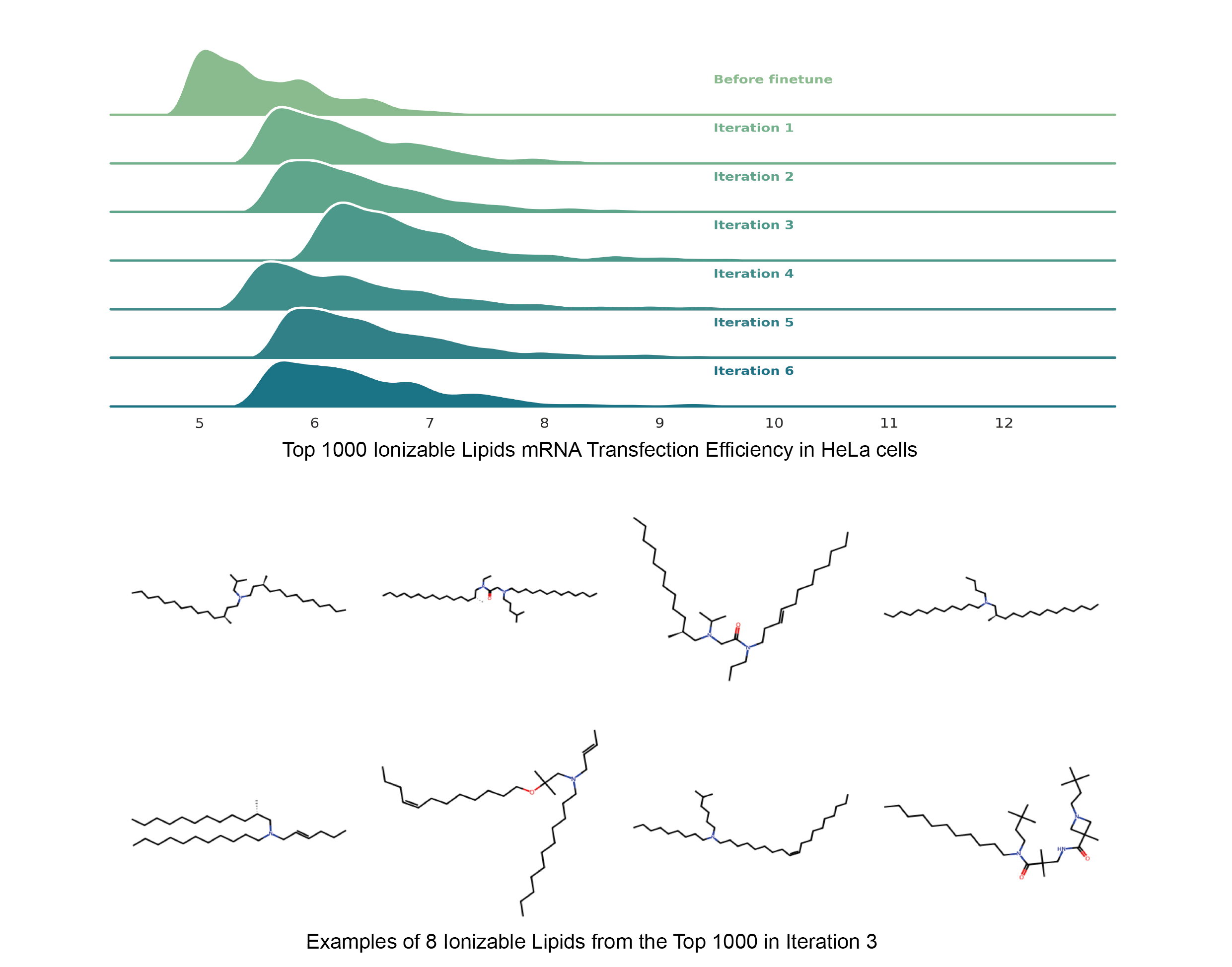}
    \caption{\textbf{Ionizable Lipids mRNA Transfection Efficiency in HeLa Cells.}}
    \label{fig:optimizatiion}
\end{figure}
We have developed an ionizable lipid version of Synthesis-DAGs. Our next goal is to optimize these lipids to enhance mRNA transfection efficiency in specific target cells. In Bradshaw’s approach \citep{synthesisdags}, the Synthesis-DAGs model is capable of iterative optimization to generate optimal novel molecules along their synthesis pathways. For this experiment, we utilized the AGILE prediction model, which estimates mRNA transfection efficiency in HeLa cells \citep{agile2}. To ensure compatibility with AGILE, we constrained our generator to produce only two-tail lipids, consistent with the lipids studied in AGILE's research. Additionally, to improve lipid stability, we restricted lipid tail lengths to 10 carbons or longer. In each iteration, we sampled 5,000 new DAGs from the model and selected the top 1,000 DAGs based on predicted transfection efficiency. The model was then fine-tuned for two rounds using these selected DAGs.
Figure \ref{fig:optimizatiion} shows the distribution of the top 1000 transfection efficiency scores for ionizable lipids sampled from the model before and after 1–6 fine-tuning iterations. Before iteration 4, we observe an increasing trend in transfection efficiency as fine-tuning progresses. This shift demonstrates the ability of our method to identify optimal lipid structures for mRNA delivery to HeLa cells while maintaining synthesizability. However, continuing the iterative fine-tuning process does not guarantee a consistently increasing transfection efficiency, as a decline is observed between iteration 4 and iteration 3. Figure \ref{fig:optimizatiion}
also includes 8 examples of ionizable lipids with the highest transfection efficiency sampled during iteration 3.

\section{Conclusion} In this work, we constructed a comprehensive ionizable lipid synthesis dataset from ZINC's synthetically accessible compounds, consisting of over 70,000 synthesis paths. Using this dataset, we developed a deep generative model that achieved an 83.4\% success rate in generating structurally diverse and synthesizable ionizable lipids, complete with synthesis paths. We also demonstrated the potential to identify ionizable lipids with high mRNA transfection efficiency in target cells.

This study highlights the application of deep generative models in ionizable lipid synthesis, combining advanced computational techniques with specific chemical synthesis challenges. Our approach efficiently generates synthesizable ionizable lipids, showing promise for advancing lipid-based RNA delivery systems.

However, it is crucial to recognize the limitations of this study. First, the validity of the predictors, including the property predictors and the reaction predictor, directly impacts the reliability of the generated synthesis DAGs. Second, the validity of the proposed synthesis pathways has not been evaluated in this work. We aim to address these issues in future research.

In future work, we will collaborate with organic chemists to synthesize the most promising ionizable lipids identified by our model and validate their mRNA transfection efficiency in wet lab experiments. This will contribute to building a more comprehensive ionizable lipid library to support the design of mRNA delivery systems.
\begin{ack}
The authors express their gratitude to Asal Mehradfar and Mohammad Shahab Sepehri for curating the lipid dataset utilized in training the lipid classifier and for developing the Lipid Analyzer toolkit.
\end{ack}

\bibliographystyle{plainnat}
\bibliography{ref}
\newpage
\section*{NeurIPS Paper Checklist}

\begin{enumerate}

\item {\bf Claims}
    \item[] Question: Do the main claims made in the abstract and introduction accurately reflect the paper's contributions and scope?
    \item[] Answer: \answerYes{} % Replace by \answerYes{}, \answerNo{}, or \answerNA{}.
    \item[] Justification: The paper's contribution and scope are clearly addressed in the main claim made in the abstract and introduction.

\item {\bf Limitations}
    \item[] Question: Does the paper discuss the limitations of the work performed by the authors?
    \item[] Answer: \answerYes{} % Replace by \answerYes{}, \answerNo{}, or \answerNA{}.
    \item[] Justification: The limitations are addressed in conclusion section.

\item {\bf Theory Assumptions and Proofs}
    \item[] Question: For each theoretical result, does the paper provide the full set of assumptions and a complete (and correct) proof?
    \item[] Answer: \answerNA{} % Replace by \answerYes{}, \answerNo{}, or \answerNA{}.
    \item[] Justification: The paper does not include theoretical results.

    \item {\bf Experimental Result Reproducibility}
    \item[] Question: Does the paper fully disclose all the information needed to reproduce the main experimental results of the paper to the extent that it affects the main claims and/or conclusions of the paper (regardless of whether the code and data are provided or not)?
    \item[] Answer: \answerYes{} % Replace by \answerYes{}, \answerNo{}, or \answerNA{}.
    \item[] Justification: All the details of dataset construction and experiment details are listed.

\item {\bf Open access to data and code}
    \item[] Question: Does the paper provide open access to the data and code, with sufficient instructions to faithfully reproduce the main experimental results, as described in supplemental material?
    \item[] Answer: \answerYes{} % Replace by \answerYes{}, \answerNo{}, or \answerNA{}.
    \item[] Justification: All the source code are publicly available.

\item {\bf Experimental Setting/Details}
    \item[] Question: Does the paper specify all the training and test details (e.g., data splits, hyperparameters, how they were chosen, type of optimizer, etc.) necessary to understand the results?
    \item[] Answer: \answerYes{} % Replace by \answerYes{}, \answerNo{}, or \answerNA{}.
    \item[] Justification: All details are included.

\item {\bf Experiment Statistical Significance}
    \item[] Question: Does the paper report error bars suitably and correctly defined or other appropriate information about the statistical significance of the experiments?
    \item[] Answer: \answerNo{} % Replace by \answerYes{}, \answerNo{}, or \answerNA{}.
    \item[] Justification: There is no error bars due to computational cost.

\item {\bf Experiments Compute Resources}
    \item[] Question: For each experiment, does the paper provide sufficient information on the computer resources (type of compute workers, memory, time of execution) needed to reproduce the experiments?
    \item[] Answer:  \answerYes{} % Replace by \answerYes{}, \answerNo{}, or \answerNA{}.
    \item[] Justification: The compute resource is address in experiment implementations section.
    
\item {\bf Code Of Ethics}
    \item[] Question: Does the research conducted in the paper conform, in every respect, with the NeurIPS Code of Ethics \url{https://neurips.cc/public/EthicsGuidelines}?
    \item[] Answer: \answerYes{} % Replace by \answerYes{}, \answerNo{}, or \answerNA{}.
    \item[] Justification: We confirm that we followed, in every respect, with the NeurIPS Code of Ethics.

\item {\bf Broader Impacts}
    \item[] Question: Does the paper discuss both potential positive societal impacts and negative societal impacts of the work performed?
    \item[] Answer: \answerYes{} % Replace by \answerYes{}, \answerNo{}, or \answerNA{}.
    \item[] Justification: Addressed in abstract, introduction and conclusion.
    
\item {\bf Safeguards}
    \item[] Question: Does the paper describe safeguards that have been put in place for responsible release of data or models that have a high risk for misuse (e.g., pretrained language models, image generators, or scraped datasets)?
    \item[] Answer: \answerNA{} % Replace by \answerYes{}, \answerNo{}, or \answerNA{}.
    \item[] Justification: The paper poses no such risks.

\item {\bf Licenses for existing assets}
    \item[] Question: Are the creators or original owners of assets (e.g., code, data, models), used in the paper, properly credited and are the license and terms of use explicitly mentioned and properly respected?
    \item[] Answer: \answerYes{} % Replace by \answerYes{}, \answerNo{}, or \answerNA{}.
    \item[] Justification:  All source codes, models, and dataset we used are properly cited.

\item {\bf New Assets}
    \item[] Question: Are new assets introduced in the paper well documented and is the documentation provided alongside the assets?
    \item[] Answer: \answerYes{} % Replace by \answerYes{}, \answerNo{}, or \answerNA{}.
    \item[] Justification: All source codes of the paper are publicly available and well documented.

\item {\bf Crowdsourcing and Research with Human Subjects}
    \item[] Question: For crowdsourcing experiments and research with human subjects, does the paper include the full text of instructions given to participants and screenshots, if applicable, as well as details about compensation (if any)? 
    \item[] Answer: \answerNA{} % Replace by \answerYes{}, \answerNo{}, or \answerNA{}.
    \item[] Justification: The paper does not involve crowdsourcing nor research with human subjects.

\item {\bf Institutional Review Board (IRB) Approvals or Equivalent for Research with Human Subjects}
    \item[] Question: Does the paper describe potential risks incurred by study participants, whether such risks were disclosed to the subjects, and whether Institutional Review Board (IRB) approvals (or an equivalent approval/review based on the requirements of your country or institution) were obtained?
    \item[] Answer: \answerNA{} % Replace by \answerYes{}, \answerNo{}, or \answerNA{}.
    \item[] Justification: The paper does not involve crowdsourcing nor research with human subjects.

\end{enumerate}

\end{document}